\documentclass{article} 
\usepackage{nips15submit_e,times}
\usepackage{hyperref}
\usepackage{url}
\usepackage{amsmath}
\usepackage{placeins}
\usepackage{graphicx}

\title{Fantasy Football Prediction}
\author{
Roman Lutz\\
College of Information and Computer Sciences\\
University of Massachusetts Amherst\\
Amherst, MA 01003 \\
\texttt{romanlutz@cs.umass.edu} 
}
\nipsfinalcopy 

\DeclareMathOperator*{\argmin}{arg\,min}

\begin{document}

\maketitle


\section{Introduction}
The ubiquity of professional sports and specifically the NFL have lead to an increase in popularity for Fantasy Football. Every week, millions of sports fans participate in their Fantasy Leagues. The main tasks for users are the draft, the round-based selection of players before each season, and setting the weekly line-up for their team. For the latter, users have many tools at their disposal: statistics, predictions, rankings of experts and even recommendations of peers. There are issues with all of these, though. Most users do not want to spend time reading statistics. The prediction of Fantasy Football has barely been studied and are fairly inaccurate. The experts judge mainly based on personal preferences instead of unbiased measurables. Finally, there are only few peers voting on line-up decisions such that the results are not representative of the general opinion. Especially since many people pay money to play, the prediction tools should be enhanced as they provide unbiased and easy-to-use assistance for users.
This paper provides and discusses approaches to predict Fantasy Football scores of Quarterbacks with relatively limited data. In addition to that, it includes several suggestions on how the data could be enhanced to achieve better results. The dataset consists only of game data from the last six NFL seasons. I used two different methods to predict the Fantasy Football scores of NFL players: Support Vector Regression (SVR) and Neural Networks. The results of both are promising given the limited data that was used.
After an overview of related work in Section~\ref{sec:rel}, I present my solution. Afterwards, I describe the data set in greater detail before Section~\ref{sec:exp} explains the experiments and show the results. Finally, Section~\ref{sec:disc} discusses the findings and possible future work.
\section{Related Work}
\label{sec:rel}
Most research in sports prediction focuses on predicting the winner of a match instead of Fantasy Football scores or specific game stats that are important for Fantasy Football. Min et al.~[5] used Bayesian inference and rule-based reasoning to predict the result of American Football matches. Their work is based on the idea that sports is both highly probabilistic and at the same time based on rules because of team strategies. Sierra, Fosco and Fierro~[1] used classification methods to predict the outcome of an American Football match based on a limited number of game statistics excluding most scoring categories. In their experiments, linear Support Vector Machines had by far the most accurate predictions and ranked above all expert predictions they compared with. Similarly, Harville~[8] used a linear model for the same task. A Neural Network approach to predicting the winner of College Football games was proposed by Pardee~[7]. His accuracy of $76\%$ improved on expert predictions. Stefani~[9] used an improved least squares rating system to predict the winner for nearly 9000 games from College and Pro Football and even other sports.
Fokoue and Foehrenbach~[2] have analyzed important factors for NFL teams with Data Mining. Their results are especially useful for this work because the first step for predicting Fantasy Football scores involves identifying important features. Even though Spann and Skiera's work on sports predictions~[3] is unrelated to American Football, their prediction markets could be used for Fantasy Football predictions. The prices on the prediction market determine the probability of a specific outcome. Such prediction markets can be used for various purposes, for example, in companies such as Google~[4]. In his blog on sports analysis, Rudy~[6] used regression models to predict Fantasy Football scores. His findings suggest that modelling positions separately improves the accuracy of a model. 


\section{Proposed Solution}
A large part of my work is to create a proper dataset from the real game data. The raw game data has to be filtered and manipulated such that only relevant data cases are used later. The predictions will be made with two different methods: Support Vector Regression (SVR) and Neural Networks. Linear models like SVR have been very successful for predicting the winner of a match~[1,8]. Neural Networks are able to adjust to the data and therefore especially useful when the structure of the data is not known. In this specific case, I have no prior knowledge of whether linear models perform well or not, so the Neural Networks will at least provide another prediction to compare the results of SVR with. As the two approaches do not have much in common, I will describe them separately.
\subsection{Support Vector Regression}
Support Vector Regression (SVR) is a linear regression model. Compared to other models, SVR is $\epsilon$-insensitive, i.e. when fitting the model to data the error is not determined on a continuous function. Instead, deviation from the desired target value $y_i$ is not counted within $\epsilon$ of the target value $y_i$. More specifically, SVR uses a regression function $f_{SVR}$ with feature vectors $x_i$ and their labels $y_i$ as follows:
\[
f_{SVR}(x) = \left(\sum_{d=1}^D w_dx_d\right) + b = xw + b
\]
where $w$ and $b$ are chosen such that
\[
w^*,b^* = \argmin_{w,b} \dfrac{C}{N} \sum_{i=1}^N V_\epsilon(y_i - (x_iw+b)) + \|w\|_2^2
\]
$V_\epsilon$ is a function that returns $0$ if the absolute value of its argument is less than $\epsilon$ and otherwise calculates the difference of the absolute value of its argument and $\epsilon$. Therefore, the loss increases linearly outside the $\epsilon$-insensitive corridor. $C$ is a regularization parameter and chosen from $(0,1]$. The smaller $C$ is chosen, the less influence is given to the features. The use of a kernel enables SVR to work well even when the structure of the data is not suitable for a linear model. There are several options for kernels which will be examined in Section~\ref{sec:exp}. The hyperparameter $\gamma$ for some kernels describes how close different data cases have to be in order to be identified as similar by the model.

\subsection{Neural Networks}
Neural Networks have the advantage that they adapt to the problem by learning the characteristics of the data set. After an initial input of the features, there are potentially multiple hidden layers and finally the output layer. In each layer there are so-called neurons or hidden units which perform computations. From layer to layer, there are connections such that the outputs of the previous layer's neurons are the inputs of the next layer's neurons. In order to work for regression the Neural Net is configured to have a linear output layer. The hidden layer units should not be linear. Common choices for the activation function are hyperbolic tangent ($tanh$) or sigmoid. 
In my experiment, I only used Neural Networks with one hidden layer. The activation function in the hidden layer for the $k$th unit is
\[h_k = \dfrac{1}{1+ \exp\left(-\left(\sum\limits_{d=1}^D w_{dk} + b_{dk}\right)\right)}\]
While the hidden layers can be non-linear, the output layer has to consist of a linear function:
\[
\hat{y} = \sum_k w^o_kh_k + b^o
\]
The parameters $w^o_k$, $b^o$ for the output layer and $w_k$, $b_k$ for the hidden layer are learnt over multiple epochs with Backpropagation. The data cases are evaluated on the Neural Network and the error is determined. Then, updates are performed based on the contribution of each hidden unit to the error, which is calculated by using derivatives going backwards through the Network.

\subsection{Pipeline}
The Neural Networks are simply given the data set in the original form. Feature selection and normalization are not performed. The Neural Network can implicitly do this or not depending on whether it improves their predictions.
Before applying SVR, I scaled the features down to the interval $[0,1]$ in order to improve the performance specifically for linear and polynomial kernel SVR. After the normalization comes the feature selection. There are three options: no feature selection, manual feature selection and Recursive Feature Elimination with Cross Validation (RFECV). All of them have certain advantages and problems. Not using feature selection can result in inaccurate predictions because of correlated features. Manual feature selection requires domain knowledge. Lastly, RFECV takes a lot of time depending on the number of features, but the results should be reasonably well since the elimination is cross validated. After the feature selection the hyperparameters for SVR have to be determined. A reasonable number of configurations is tested several times on a held-out validation set after being trained on the rest of the training data. The hyperparameter configuration with the best average score is finally selected and used for all further operations. This includes fitting the model to the whole training data set and finally predicting the Fantasy Football scores of the test cases.

\section{Data Set}
The data set consists of NFL game data from 2009 to 2014. I accessed it with the API from \href{https://github.com/BurntSushi/nflgame}{github.com/BurntSushi/nflgame} which gets the data from \href{NFL.com}{NFL.com}. Before using it to make predictions I performed several operations. First of all, I filtered the data such that only Quarterbacks (QB) with at least $5$ passes are selected. This restriction is necessary such that non-QB players or backup QBs are not taken in to account. Then, for every game I included as features the current age of the QB, his experience in years as a professional, the stats of the previous game, the average stats of the last $10$ games as well as the stats of the opposing defense in their last game and their average over the last $10$ games. The stats for a QB include $12$ features that show the performance in passing and rushing as well as turnovers. The values are all treated as continuous real values. For defenses, there are $4$ categories, namely the number of points allowed, passing and rushing yards allowed as well as turnovers forced. The target value in each case is the actual Fantasy Football score the QB received for the given game. I used the NFL's standard scoring system which is described on \href{http://www.nfl.com/fantasyfootball/help/nfl-scoringsettings}{nfl.com/fantasyfootball/help/nfl-scoringsettings}.
In order to have sigificant past data even for the first data cases, I did not use the first year, 2009. I split the data into training and test data such that the seasons 2010 to 2013 belong to the training set and the 2014 season is the test data. As a result, there are 2167 training cases and 553 test cases.
First-year players become a separate problem because the predictions can not be based on their past production. To overcome this, they are assigned the average over all first-year QB per-game average stats for the first game. From the second game on, their own statistics are used. 
Even though the data access through the API is limited to years 2009-2014, this is not necessarily a limitation. As various independent statistics and reports~[11,12,13,14] show, Football has evolved especially over the last few years. Such changes also influence Fantasy Football. As a consequence, the data from ten years ago might not properly represent today's games any more thus affecting the predictions negatively.
The test data includes lots of cases with QBs that would never be used in Fantasy Football because of a lack of experience, production or inconsistency. Therefore it makes sense to restrict the evaluation to the best QBs that actually have a chance to be used in Fantasy Football. In standard leagues with $12$ teams one QB starts for every team, so the evaluation considers the predictions of the $24$ best QBs (see Appendix List~\ref{listof24qbs}).

\section{Experiments and Results}
\label{sec:exp}
\begin{figure}
\includegraphics[scale=0.35]{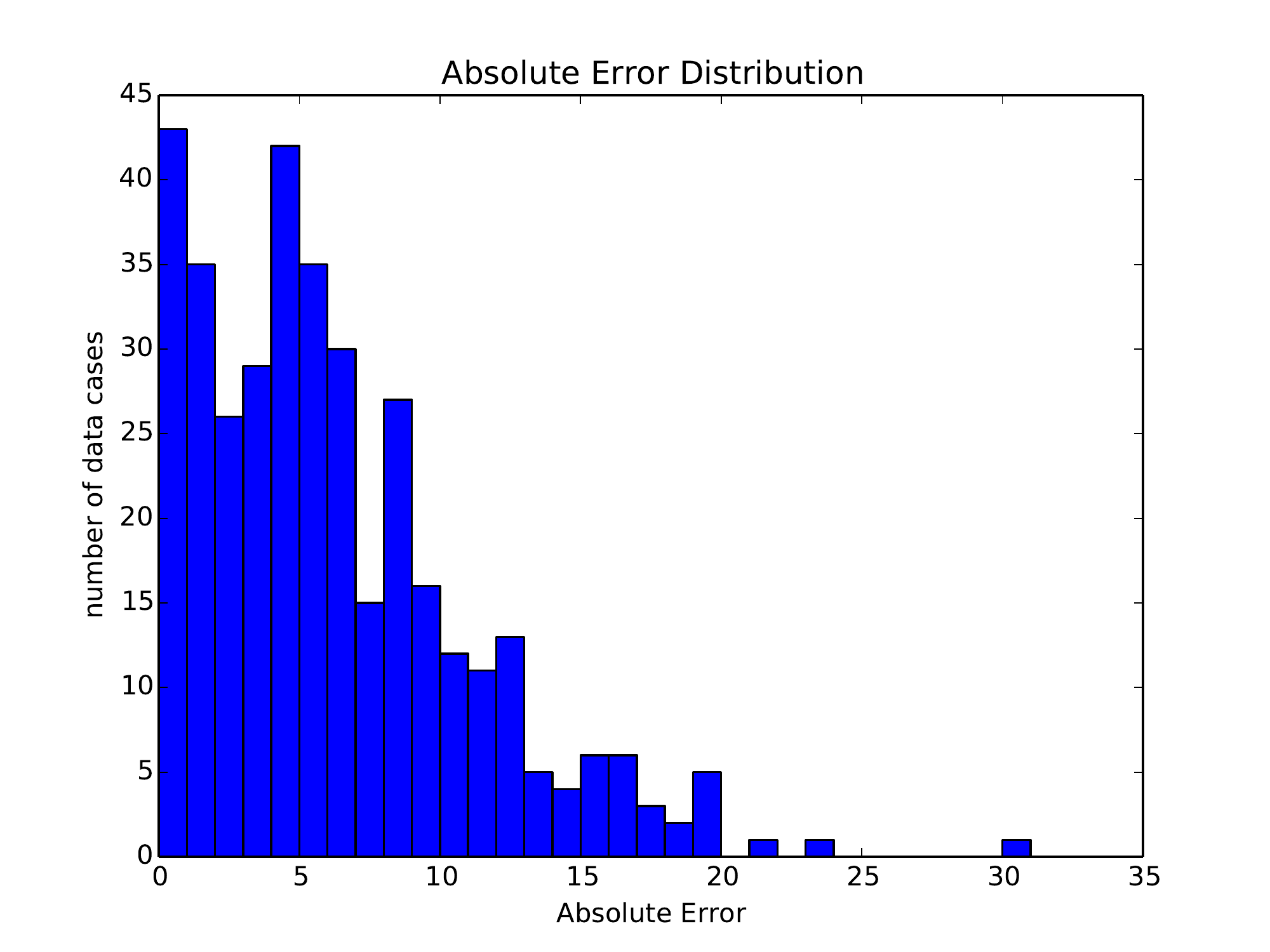}
\includegraphics[scale=0.35]{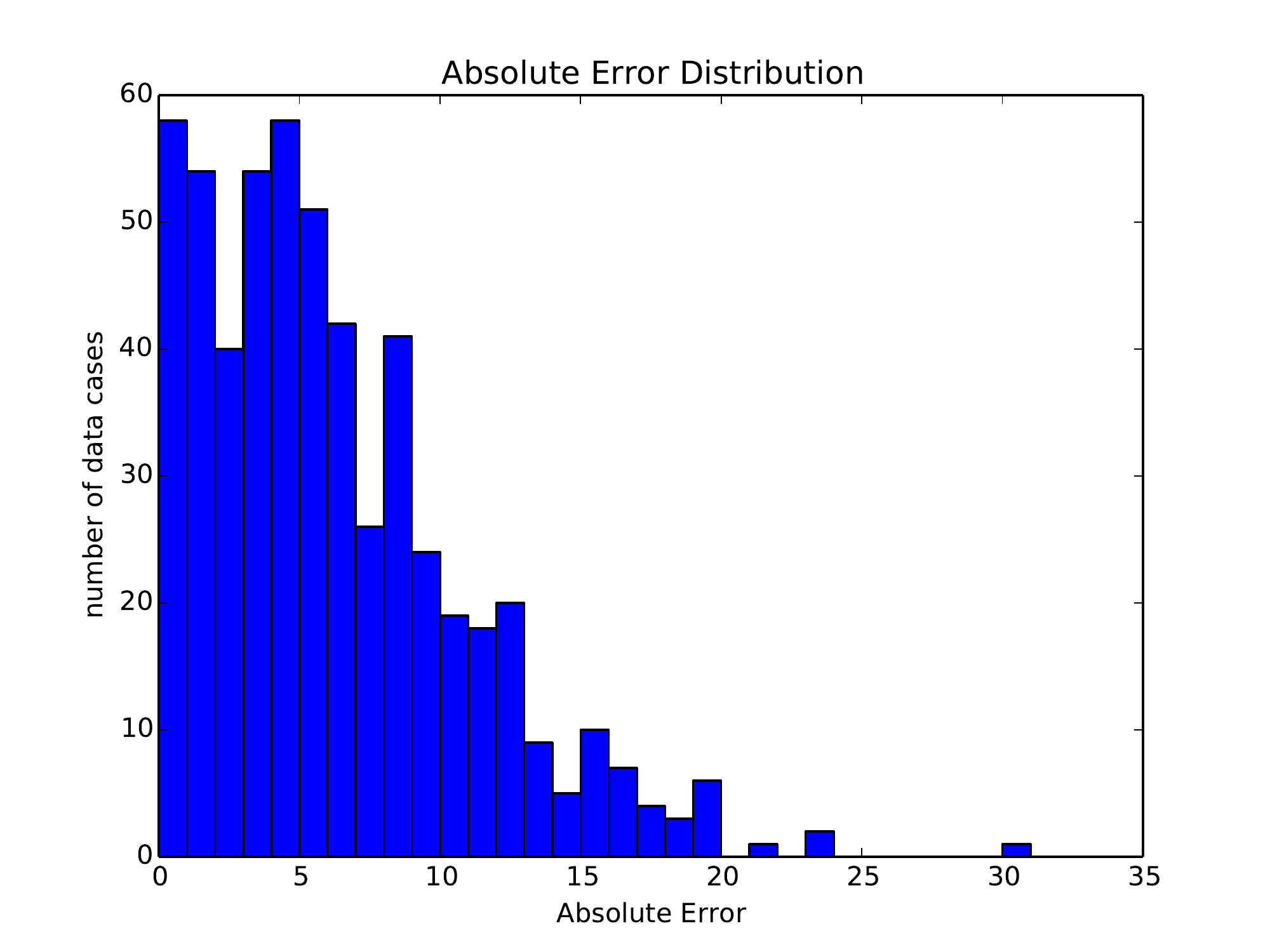}
\caption{The figures show the distribution of the absolute errors of the predictions of SVR with no feature selection and hyperparameters set to $C=0.25$, $\epsilon=0.25$, $\text{kernel}=\text{linear}$. \textbf{Left:} Absolute error over all test cases. \textbf{Right:} Absolute error over test cases involving $24$ best players.}
\label{fig:abserrordist}
\end{figure}

\begin{table}
\centering
\begin{tabular}{lllllll}\hline
Feature Sel. & RMSE (all) & RMSE (24) & MAE (all) & MAE (24) & MRE (all) & MRE (24) \\ \hline
None & $7.815$ & $7.925$ & $6.238$ & $6.265$ & $0.453$ & $0.419$ \\
RFECV & $7.759$ & $7.833$ & $6.221$ & $6.248$ & $0.448$ & $0.418$ \\
manual & $7.796$ & $7.914$ & $6.224$ & $6.256$ & $0.450$ & $0.418$ \\
\hline
\end{tabular}
\caption{The results of applying SVR after no feature selection, RFECV and manual feature selection. The hyperparameters were set to $C=0.25$, $\epsilon=0.25$, $\text{kernel}=\text{linear}$. For reasons of comparability with other sources three different errors are shown: Root Mean Squared Error (RMSE), Mean Absolute Error (MAE) and Mean Relative Error (MRE). MRE is defined as $\frac{|y-\text{prediction}|}{\text{prediction}}$. All errors are shown for the whole test set (all) and for data cases which involved the best $24$ players only.}
\label{tab:SVR}
\end{table}
In this section, I will talk about the different experiments and the obtained results. First of all, I tried Support Vector Regression (SVR) using the scikit-learn implementation~[15]. There are several features that probably do not influence the result very much, e.g. the number of two point conversions in the last game, so feature selection could actually improve the accuracy. The method I chose for feature selection is \textit{Recursive Feature Elimination with Cross Validation} (RFECV). It recursively eliminates features and checks if the regression method's results improve by cross validating. In order to reduce the running time I applied the feature selection before the hyperparameter selection. The assumption was that no matter which regression method and hyperparameters are chosen, the important features will always be more or less the same. The selected features were the age, the number of years as a professional player, the number of passing attempts and the number of successful passing two-point conversions (both from last 10 games). This is a very small subset of the features since important features like the touchdowns were not taken into consideration and rarely occurring, low-weighted features like two-point conversion stats are included. Therefore, I also tried two other ways: no feature selection and manual feature selection. For the manual feature selection, I simply removed the two-point conversion stats. The results are compared in Figure~\ref{tab:SVR}. Interestingly, the SVR with RFECV performed best, followed by SVR with manual feature selection and SVR without feature selection. The reason for this is most likely that the selected hyperparameters were $C=0.25$, $\epsilon=0.25$ and a linear kernel. Having a small regularization value $C$ means that the influence of the feature values on the prediction is reduced.
Even though the feature normalization did not influence the error, it sped up the running time immensely. This allowed for more configurations in the hyperparameter selection for SVR. All possible combinations of the following values were used:
\begin{align*}&\text{kernel} \in \{\text{radial basis function},\text{sigmoid}, \text{linear}, \text{polynomial}\}, \\&C \in \{0.25, 0.5, 0.75, 1.0\}, \quad\epsilon \in \{0.05, 0.1, 0.15, 0.2, 0.25\}, \\ &\gamma \in \{0, 0.05, 0.1, 0.15\}, \quad\text{degree}\in\{2,3\}\end{align*}
As already mentioned above, the best configuration was $C=0.25$, $\epsilon=0.25$ and $\text{kernel}=\text{linear}$, although it is worth mentioning that the difference is very small. $\gamma$ and the degree are only used for some of the kernels.
Figure~\ref{fig:abserrordist} shows both the distribution of the absolute error in the whole test set and when considering only the cases with the best $24$ players. The overall distribution is quite similar which is also represented in the numbers in Table~\ref{tab:SVR}. Therefore, predictions of all QBs seem to roughly as difficult as predictions of the top $24$ QBs only.

Secondly, I used Neural Networks based on the PyBrain library~[16]. In order to determine the right number of epochs $n_{epochs}$, number of hidden units $n_{hidden}$ and the type of neurons in the hidden layer $t_{hidden}$, every combination of the following values was used:
\[n_{epochs} \in \{10,50,100,1000\}, \quad n_{hidden} \in \{10,25,50,100\}, \quad t_{hidden} \in \{\text{Sigmoid}, \text{Tanh}\}\]
The results are shown in the Appendix in Table~\ref{tab:neuralnetresult}. In comparison with the best SVR results, most Neural Nets performed worse on both on the whole test set and for only the $24$ best players in all error categories. The only configuration that achieved significantly better results than all others was $n_{epochs} = 50$, $n_{hidden} = 50$ and $t_{hidden}=\text{Sigmoid}$. Its RMSE, MAE and MRE were $7.868$, $6.235$ and $0.413$ on the cases involving only the $24$ best players. Interestingly, the errors for the best $24$ players were much lower than for the whole test set. The MAE and MRE are actually better than the best results of the SVRs.

\section{Discussion and Conclusions}
\label{sec:disc}
All in all, the errors were still very high. For example, the MAE of the best prediction was more than $6$ points, which can make a difference in most Fantasy Football games. Considering that I only predicted the scores for one position, Quarterback, there are still several more positions on a team thus potentially increasing the overall error. The comparison with other sources of predictions is hard because most websites just use experts. The ones that actually project with a model, e.g. ESPN, do not openly write about their accuracy. Reda and Stringer~[10] analyzed ESPN's accuracy, but they used multiple positions at once such that I could not compare it properly. It is encouraging, though, that their MAE histograms show the same Bell Curve shape and have approximately the same variance.
Apart from that, there are a few interesting observations from this experiment. The feature selection with RFECV selected only $4$ features. I assume that many of the features are correlated and therefore do not add much extra value to the predictions. The only ways the accuracy can be significantly improved is by adding new features or by enhancing existing ones. As the feature selection indicated, having both last week's statistics and the average over the last ten games does not provide much extra information. In order to take the trend better into account, the exponentially weighted moving average (EWMA) could be used to substitute the current game statistics. The EWMA is calculated as
\[
S_t = \alpha G_t + (1-\alpha) S_{t-1}
\]
where $S_t$ is the EWMA for all games up to $t$ and $G_t$ are the games statistics for game $t$. This could be done for the last few games or even over the whole career.\\
There are also several other interesting features that could be taken into account, such as the injury report status, suspensions, draft position and college statistics for first-year players, postseason and preseason performance, overall team performance and changes on the team such as trades or injuries. One could even go further and analyze Twitter posts on a specific player or factor in expert rankings and predictions from other sources. Most of these are not accessible with the used API and exceed the scope of this project, but seem promising for future work.
So far, only Neural Networks and Support Vector Regression were used. Other models are conceivable for this task, too. But specifically Neural Networks with multiple hidden layers could be useful since the Networks with only one layer performed already quite well.
Overall, as the total number of users of Fantasy Football Leagues and the amounts of invested money increase the demand for accurate predictions will grow and probably lead to more research on the topic.

\subsubsection*{References}

\small{

[1] Sierra, A., Fosco, J., Fierro, C., \& Tiger, V. T. S. Football Futures. Retrieved April 30, 2015, from http://cs229.stanford.edu/proj2011/SierraFoscoFierro-FootballFutures.pdf.

[2] Fokoue, E., \& Foehrenbach, D. (2013). A Statistical Data Mining Approach to Determining the Factors that Distinguish Championship Caliber Teams in the National Football League.

[3] Spann, M., \& Skiera, B. (2009). Sports forecasting: a comparison of the forecast accuracy of prediction markets, betting odds and tipsters. Journal of Forecasting, 28(1), 55-72.

[4] Cowgill, B. (2015). Putting crowd wisdom to work. Retrieved April 30, 2015, from http://googleblog.blogspot.com/2005/09/putting-crowd-wisdom-to-work.html

[5] Min, B., Kim, J., Choe, C., Eom, H., \& McKay, R. B. (2008). A compound framework for sports results prediction: A football case study. Knowledge-Based Systems, 21(7), 551-562.

[6] Rudy, K. (n.d.). The Minitab Blog. Retrieved April 27, 2015, from http://blog.minitab.com/blog/the-statistics-game

[7] Pardee, M. (1999). An artificial neural network approach to college football prediction and ranking. University of Wisconsin–Electrical and Computer Engineering Department.

[8] Harville, D. (1980). Predictions for National Football League games via linear-model methodology. Journal of the American Statistical Association, 75(371), 516-524.

[9] Stefani, R. T. (1980). Improved least squares football, basketball, and soccer predictions. IEEE transactions on systems, man, and cybernetics, 10(2), 116-123.

[10] Reda, G., \& Stringer, M. (2014). Are ESPN's fantasy football projections accurate? Retrieved April 28, 2015, from http://datascopeanalytics.com/what-we-think/2014/12/09/are-espns-fantasy-football-projections-accurate

[11] Powell-Morse, A. (2013). Evolution of the NFL Offense: An Analysis of the Last 80 Years. Retrieved April 27, 2015, from http://www.besttickets.com/blog/evolution-of-nfl-offense/

[12] Wyche, S. (2012) Passing league: Explaining the NFL's aerial evolution. Retrieved April 26, 2015, from http://www.nfl.com/news/story/09000d5d82a44e69/article/passing-league-explaining-the-nfls-aerial-evolution

[13] Mize, M. (2012) The NFL is a passing league. The statistics prove it and the rules mandate it. Retrieved April 26, 2015, from http://www.thevictoryformation.com/2012/09/17/the-nfl-is-a-passing-league-the-statistics-prove-it-and-the-rules-mandate-it/

[14] Rudnitsky, M. (2013) Today’s NFL Really Is A ‘Passing League,’ As Confirmed By These Fancy Graphs That Chart The Last 80 Years Of Offense. Retrieved April 26, 2015, from http://www.sportsgrid.com/nfl/evolution-nfl-offense-charts-graphs-passing-rushing-peyton-manning/

[15] scikit-learn: Machine Learning in Python. Retrieved April 26, 2015, from http://scikit-learn.org/stable/index.html

[16] PyBrain: The Python Machine Learning Library. Retrieved April 26, 2015, from http://www.pybrain.org/

[17] Gallant, A. (2012-15) An API to retrieve and read NFL Game Center JSON data. Retrieved April 26, 2015, from https://github.com/BurntSushi/nflgame

[18] NFL.com Fantasy Football scoring settings: Retrieved April 26, 2015, from http://www.nfl.com/fantasyfootball/help/nfl-scoringsettings

}

\section*{Appendix}
\FloatBarrier
In the following, all $24$ Quarterbacks that were considered for the evaluation are listed. The selection was based on their 2014 Fantasy Football scores. For players that did not play all regular season games or were first-year players (Rookie) with no previous Fantasy Football statistics there is a note:\\
\label{listof24qbs}
Drew Brees, Ben Roethlisberger, Andrew Luck, Peyton Manning, Matt Ryan, Eli Manning, Aaron Rodgers, Philip Rivers, Matthew Stafford, Tom Brady, Ryan Tannehill, Joe Flacco, Jay Cutler (15 games), Tony Romo (15 games), Russell Wilson, Andy Dalton, Colin Kaepernick, Brian Hoyer (14 games, 13 starts), Derek Carr (Rookie), Alex Smith (15 games), Cam Newton (14 games), Kyle Orton, Teddy Bridgewater (13 games, 12 starts, Rookie), Blake Bortles (14 games, 13 starts, Rookie)

\begin{table}
\centering
\caption{The results from the Neural Networks on the Fantasy Football predictions. Every run is defined by the number of epochs $n_{epochs}$, the number of hidden units $n_{hidden}$ and the type of neurons in the hidden unit $t_{hidden}$. The results include the Root Mean Squared Error (RMSE), Mean Absolute Error (MAE) and Mean Relative Error (MRE) on the whole test data as well as the RMSE, MAE and MRE for the best $24$ players.}
\label{tab:neuralnetresult}
\vspace*{0.5cm}
\hspace*{-0.7cm}
\begin{tabular}{lllllllll}\hline
$n_{epochs}$ & $n_{hidden}$& $t_{hidden}$ & RMSE(all)& MAE(all) & MRE(all) & RMSE(24) & MAE(24) & MRE(24)\\\hline
10 &10 &Sigm & 8.060258 & 6.391866 &  0.461691 & 8.142453 & 6.381999 & 0.460978 \\ 
10 &10 &Tanh  &8.058488 & 6.391081 &  0.460524 & 8.132701 & 6.375831 & 0.459425 \\ 
10 &25 &Sigm & 8.049170 & 6.387444 &  0.457333 & 8.105430 & 6.358798 & 0.455058 \\ 
10 &25 &Tanh  &8.069334 & 6.396201 &  0.466794 & 8.185348 & 6.409744 & 0.467782 \\ 
10 &50 &Sigm & 8.054337 & 6.387369 &  0.472183 & 8.222789 & 6.433377 & 0.473597 \\ 
10 &50 &Tanh  &8.057394 & 6.390635 &  0.459761 & 8.126329 & 6.371757 & 0.458403 \\ 
10 &100 &Sigm & 8.127506 & 6.424212 &  0.488398 & 8.371598 & 6.526674 & 0.496181 \\ 
10 &100 &Tanh  &8.182758 & 6.450737 &  0.502695 & 8.499078 & 6.602691 & 0.514691 \\ 
50 &10 &Sigm & 8.059963 & 6.391739 &  0.461502 & 8.140871 & 6.381004 & 0.460727 \\ 
50 &10 &Tanh  &8.103019 & 6.412570 &  0.480483 & 8.302668 & 6.484871 & 0.485900 \\ 
50 &25 &Sigm & 8.049401 & 6.389763 &  0.451924 & 8.060769 & 6.331248 & 0.447786 \\ 
50 &25 &Tanh  &8.059079 & 6.391350 &  0.460923 & 8.136031 & 6.377946 & 0.459957 \\ 
50 &50 &Sigm & 8.061196 & 6.441132 &  0.429238 & 7.867709 & 6.235217 & 0.413366 \\ 
50 &50 &Tanh  &8.049278 & 6.390949 &  0.450475 & 8.048565 & 6.323527 & 0.445747 \\ 
50 &100 &Sigm & 8.054849 & 6.385918 &  0.463524 & 8.157339 & 6.390617 & 0.463520 \\ 
50 &100 &Tanh  &8.071710 & 6.392687 &  0.448577 & 8.060837 & 6.323155 & 0.444378 \\ 
100 &10 &Sigm & 8.087679 & 6.404901 &  0.474837 & 8.254096 & 6.453887 & 0.478469 \\ 
100 &10 &Tanh  &8.048337 & 6.393241 &  0.445171 & 8.005743 & 6.296778 & 0.438454 \\ 
100 &25 &Sigm & 8.079818 & 6.411072 &  0.470735 & 8.199887 & 6.418849 & 0.470043 \\ 
100 &25 &Tanh  &8.061363 & 6.383954 &  0.463902 & 8.167160 & 6.398136 & 0.464922 \\ 
100 &50 &Sigm & 8.060657 & 6.393333 &  0.460352 & 8.133239 & 6.377483 & 0.459263 \\ 
100 &50 &Tanh  &8.038246 & 6.381052 &  0.446976 & 8.014688 & 6.297888 & 0.441000 \\ 
100 &100 &Sigm & 8.069325 & 6.404321 &  0.464389 & 8.149586 & 6.386675 & 0.462122 \\ 
100 &100 &Tanh  &8.069874 & 6.408302 &  0.459283 & 8.118948 & 6.383348 & 0.457499 \\ 
1000 &10 &Sigm & 8.051943 & 6.389734 &  0.455206 & 8.087935 & 6.348112 & 0.452241 \\ 
1000 &10 &Tanh  &8.056874 & 6.388611 &  0.450233 & 8.036326 & 6.302215 & 0.444779 \\ 
1000 &25 &Sigm & 8.066095 & 6.396302 &  0.467444 & 8.188647 & 6.411823 & 0.468297 \\ 
1000 &25 &Tanh  &8.053281 & 6.389806 &  0.456502 & 8.098831 & 6.354800 & 0.454001 \\ 
1000 &50 &Sigm & 8.055530 & 6.381625 &  0.460907 & 8.134901 & 6.366414 & 0.459836 \\ 
1000 &50 &Tanh  &8.073334 & 6.392310 &  0.470385 & 8.218570 & 6.435802 & 0.473622 \\ 
1000 &100 &Sigm & 8.091592 & 6.406337 &  0.479083 & 8.289962 & 6.476950 & 0.483967 \\ 
1000 &100 &Tanh  &8.049693 & 6.385986 &  0.456491 & 8.089114 & 6.348840 & 0.452432 \\ 

\end{tabular}
\end{table}
\end{document}